\documentclass[a4paper, 10pt, conference]{ieeeconf}      % Use this line for a4 paper

\IEEEoverridecommandlockouts                              % This command is only needed if 
\usepackage{graphics} % for pdf, bitmapped graphics files
\usepackage{epsfig} % for postscript graphics files
\usepackage{url}

\title{\LARGE \bf
Cloud Native Robotic Applications with GPU Sharing on Kubernetes
}

\author{Giovanni Toffetti$^{1}$, Leonardo Militano$^{1}$, Seán Murphy$^{2}$, Remo Maurer$^{1}$, and Mark Straub$^{1}$% <-this % stops a space
\thanks{This work was accepted at the IROS'22 Cloud Robotics Workshop: https://sites.google.com/view/iros22-cloud-robotics}% <-this % stops a space
\thanks{$^{1}$Zurich University of Applied Sciences (ZHAW), Switzerland
        {\tt\small [toff|milt|murm|stmr]@zhaw.ch}}%
\thanks{$^{2}$ETH Zürich, Swiss Data Science Center (SDSC), Switzerland
        {\tt\small sean.murphy@sdsc.ethz.ch}}%
}

\begin{document}

\maketitle
\thispagestyle{empty}
\pagestyle{empty}

%%%%%%%%%%%%%%%%%%%%%%%%%%%%%%%%%%%%%%%%%%%%%%%%%%%%%%%%%%%%%%%%%%%%%%%%%%%%%%%%
\begin{abstract}

In this paper we discuss our experience in teaching the Robotic Applications Programming course at ZHAW combining the use of a Kubernetes (k8s) cluster and real, heterogeneous, robotic hardware.
We discuss the main advantages of our solutions in terms of seamless ``simulation to real'' experience for students and the main shortcomings we encountered with networking and sharing GPUs to support deep learning workloads.
We describe the current and foreseen alternatives to avoid these drawbacks in future course editions and propose a more cloud-native approach to deploying multiple robotics applications on a k8s cluster.

\end{abstract}

%%%%%%%%%%%%%%%%%%%%%%%%%%%%%%%%%%%%%%%%%%%%%%%%%%%%%%%%%%%%%%%%%%%%%%%%%%%%%%%%
\section{INTRODUCTION}

The Robotic Applications Programming course (RAP) has been taught at the School of Engineering of the Zurich University of Applied Sciences since Spring 2021.
The course is offered to bachelor students in IT in their last semesters with the aim of learning how to program robotic applications using the ROS framework, as well as combining knowledge from other courses (e.g., computer vision, artificial intelligence, cloud computing) to make robots autonomous.

The course is organized in three main sections: (1) robotics (e.g.,
basic robotic HW, URDF/XACRO, rviz, poses, coordinate frames and transformations, controllers) and ROS fundamentals (communication primitives, ROS packages); (2) base capabilities (e.g., SLAM, navigation, perception, arm motion
planning and control); and (3) distributed robotic applications culminating with a yearly challenge. This year's challenge was inspired by the DARPA Subterranean Challenge\footnote{\url{https://www.subtchallenge.com}}: student groups had to program a mobile manipulator (a Robotnik Summit XL with a UR5 arm) to autonomously explore an unknown indoor area, detect coke cans, and bring them back to the starting point.

We designed the theoretical modules and practical labs of the course for students to minimize the configuration and set-up effort to work collaboratively in order to focus on software development and, at the same time, preserve the excitement of seeing one's own code running on real robotic hardware.
In order to do this we first investigated available solutions we could use for our scenario. 
We report on related work in the next section.

\section{RELATED WORK}
\label{sec:related_work}
%The interest in robotics engineering has been growing rapidly over the last few years. For hobbyists, students and professionals, the amount of robotics practitioners has steadily grown and with it the available educational content. At the same time, educational institutions at all levels are working hard to adapt their instructional programs and learning paths to integrate robotic technologies. For instance, Educational Robotics (ER) is a modern teaching practice that the teacher engages in, using robots as a tool for designing and integrating the educational process. ER was identified as an educational resource through which students acquire knowledge of different disciplines and improve their attitude and interest in STEAM disciplines (Science, Technology, Engineering, Arts and Mathematics) \cite{schina}, \cite{curto}. 

As the complexity of robotic applications is growing steadily, with the adoption of advanced solutions such as AI for semantic mapping or grasp generation, new needs appeared in terms of computation, networking and storage resources. To cope with them, Fog/Cloud-Robotic solutions are gaining traction in several domains. The possibility for remotely controlling robotic systems further reduces costs for deployment, monitoring, diagnostic and orchestration of any robotic application. This, in turn, allows for building lightweight, low cost and smarter robots as the main computation and communication burden is brought to the cloud. Since 2010, when the "cloud robotics" term was first used, several projects (e.g., RoboEarth \cite{roboearth} DAVinci \cite{davinci}) investigated the field pushing forward both research and products to appear on the market. Companies started investing in the field as they recognized the huge potential of cloud robotics. This led to open-source cloud robotics frameworks appearing in recent years. An example of this is the solution from Rapyuta Robotics\footnote{\url{https://www.rapyuta-robotics.com/}}. Similarly, commercial solutions for developers have seen the light with the big players in the Cloud field joining the run (e.g., Amazon Robomaker\footnote{\url{https://cloud.google.com/cloud-robotics/}}) or startups (e.g., Formant.io).

\subsection{Robotic Applications Development in Education}
% I don't like the next para - there are too many things in it... MOVED SOME PARTS IN THE INTRODUCTION
 Depending on the educational level and the requirements for professional knowledge of robotic application development, different teaching and learning approaches can be identified as a combination of simulation- and hardware-based solutions.

\textbf{Simulation-based learning} leverage software tools and programming languages to simulate the behavior of robots without direct interaction with a physical robot. Under this category we include web robotics as a way of learning online using a web-based platforms for simulating robots, as e.g., in \cite{alvarez}. This latter is gaining momentum with offerings such as The Construct\footnote{\url{https://www.theconstructsim.com/}} or AWS RoboMaker\footnote{\url{https://aws.amazon.com/robomaker/}} which are cloud-based simulation services that enable robotics developers to run, scale, and automate simulation without managing any infrastructure. 
% One of the most widely used simulators for ROS is Gazebo\footnote{\url{https://gazebosim.org/}} which provides 3D physics simulation for a large variety of sensors and robots. Gazebo is bundled into the full installation package of ROS, making it widely and easily available, and many robot manufacturers offer ROS packages specifically designed to support Gazebo. Other popular robotic simulators are Webots\footnote{\url{https://cyberbotics.com/}}, CoppeliaSim\footnote{\url{http://www.coppeliarobotics.com/}} and OpenRave\footnote{\url{http://www.osrobotics.org/osr/}}. Besides these, game engines are also being adapted to support robotic simulation such as, for instance, Unity\footnote{\url{https://github.com/Unity-Technologies/Unity-Robotics-Hub}}. 
Simulation based solutions are clearly useful and serve some important educational needs; however, the models on which they are based always have some limitations which can become apparent in a real world context. Further, adopting a simulation only approach does not give students experience with some of the more practical considerations associated with working with physical devices.

\textbf{Hardware-based learning} allows for direct interaction and programming of physical robots. In some simple domains and for simple applications students can safely interact directly with the hardware without necessarily first simulating the application behavior. One example of this is the LEGO® Robot Programming for kids program\footnote{\url{https://www.lego.com/en-gb/categories/coding-for-kids}} where kids build a robot, program it and interact with it; programming in this environment is based on a set of predefined tasks the robot can execute. Similar other solutions exist, but these  lack flexibility and the extensibility and customization capabilities required for real world robotics scenarios. To develop more realistic applications the use of programming languages such as Python, C++, MATLAB or frameworks like ROS is a must.  

\textbf{Hybrid learning} combines simulation and hardware-based learning where  the robotic application can be  tested in a simulated environment and deployed on the physical. In doing so we have the advantages of less costs, reduced risks of damaging expensive hardware, reduced risks of damages to third persons and things. Oftentimes, a digital copy of the robotic hardware can be used for visualization and control of the robot. In advanced solution, a digital-twin can be placed into a simulated environment while the actions and tasks are physically executed on the hardware. In this way, the simulated environment will provide inputs to the application in terms of environment (e.g., obstacles), sensing information (e.g., light, temperature), which allows to test applications in a close-to-real environment.

\section{USE CASES and REQUIREMENTS}

During the duration of the course, students are expected to use three different types of robots to familiarize themselves with different use cases and capabilities.
\begin{itemize}
    \item \textbf{Turtlebot3} (6x): these robots are used to first teach rudimentary ROS communication primitives (i.e., implementing a random-walk reading from the laser scanner and sending a cmd\_vel to move\_base), then experiment with SLAM (with gmapping and SLAM toolkit), and finally navigating waypoints (move\_base);
    \item \textbf{Niryo One Arms} (3x): these simple 6 DoF arms are used together with Realsense cameras running on Raspberry Pi4 to first learn about poses and transformations (i.e., picking a marker with a simplified script), then experimenting with MoveIt, and finally picking a random object using point cloud segmentation and the GPD\footnote{\url{https://github.com/atenpas/gpd}} library;
    \item \textbf{Summit XL}: the large mobile manipulator from Robotnik, equipped with a UR-5 arm and a Robotiq gripper, is used by students exclusively in simulation during the course labs to combine all learnt capabilities to solve the yearly challenge. The best challenge solutions are run on the physical robot at the end of the course.
\end{itemize}

In RAP, the objective is to teach students the use of ROS and application development addressing problems which typically arise in a robotics context, e.g. navigation and mapping, grasping of objects and perception. The students should be able to collaboratively develop software (in teams of 3) and quickly test it using simulation; only code working correctly in simulation is then run on physical robots. Further, \emph{embracing the Cloud Robotics paradigm}, some components of the robotic application will run on the physical robots, while others will run on the cloud (e.g., the GPD neural network which requires a GPU) or the edge (e.g., the Realsense ROS node) of the network. The objective of our system setup is that students can \emph{seamlessly transition their applications from the simulation environment to the real world context}, while not having to address the troublesome issues associated with framework setup and networking which arise in such distributed systems.

\section{SOLUTIONS and DRAWBACKS}

In this section we discuss the compute and networking environment that was used at ZHAW in the last course editions.
First of all, due to university security policy, all robots used for the course are constrained to a subnet (iot-ZHAW) that is \emph{blocked from accessing ZHAW's internal network}, where teacher and student laptops are connected. Hence, standard distributed ROS applications (requiring bidirectional TCP connections) cannot run. Unsurprisingly, this is a common restriction at many universities.

Moreover, students use their \emph{own personal laptop} to attend the course (BYOD), each with its own CPU architecture (e.g., x86 vs M1) and operating system. Installing a functioning ROS environment (Noetic) for each student was out of the question.
Virtual Machines (VMs) with preinstalled ROS are a common solution, but they require installation and management of the images, would have very different performance for each student when running simulations, and still would require some networking configuration to forward ports from the host to the VM. Finally, the isolated robot subnet would still be an issue.

We wanted students to learn from each other by \emph{working in groups} on the same codebase and simulation environment, we \emph{needed GPU-acceleration} to boost simulation performance (keeping a decent sim-to-real time ratio) as well as to run neural-network based components (e.g., the already mentioned GPD, but also instance segmentation with Mask-RCNN).

\subsection{Spring 2021: VMs and flat network}

Given the above requirements and constraints, for the first edition of the course we opted to extend our local Openstack cluster installation with a \emph{dedicated node} for the course. With 8 Nvidia Tegra GPUs per node we could allocate one GPU per VM and provide sufficient computation for 24 students (in groups of 3).
The cluster node was also running outside of the internal ZHAW network, in a subnet that could be reached by iot-ZHAW, hence bidirectional TCP communication with the robots was possible.

\begin{figure}[thpb]
  \centering      
  \includegraphics[scale=0.8]{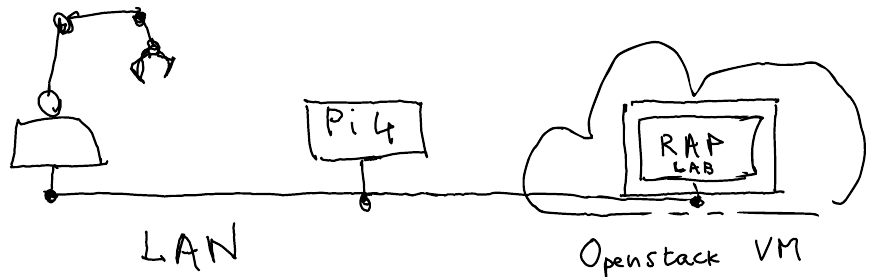}
  \caption{The robot and Pi use TCPROS to communicate with each other and with the container running the RAP lab code}
  \label{fig:vm2021}
\end{figure}

VMs were pre-instantiated by ZHAW staff, and students had a shell account. To make available the slightly different environments and components we needed for each lab, we provided students with different container images each week. They would run them with host mode networking, so that container ports would be directly addressable on the host by the robots as depicted in Figure \ref{fig:vm2021}. They could use their own laptops to access a complete ROS virtual environment through a browser with VNC.

The simulation environment would run smoothly at about 60 FPS\footnote{e.g., a short example video is available here: \url{https://www.youtube.com/watch?v=3sA1wCgaEzM}} and the steps required for transitioning from simulation to real hardware would simply be: 1) configuring the ROS\_MASTER\_URI environmental variable to match the allocated robot, 2) configuring the ROS\_HOSTNAME https://sites.google.com/view/iros22-cloud-roboticsenvironmental variable in the container to the \emph{floating IP address} of the VM. DNS name resolution of the robots / PIs would take care of the rest.

The main drawbacks of this solution were 1) the fact that we had to manage student VMs individually and they would be pinned (constantly preventing others from accessing) to GPUs; 2) students had to configure networking manually often having trouble with the concept of floating IP; 3) each VM had to be preconfigured\footnote{\url{https://github.com/gtoff/nvidia-docker-novnc}} to enable GPU acceleration of VNC with VirtualGL\footnote{\url{https://wiki.archlinux.org/title/VirtualGL#Using_VirtualGL_with_VNC}} and each container would have to be built specifically to make use of that\footnote{\url{https://github.com/icclab/rosdocked-irlab/blob/noetic/BASE_GPU/Dockerfile}}.

%TODO: maybe some considerations on latencies, storage delays, performance?

\subsection{Spring 2022: K8S and rosbridge}
In fall 2021 we set out to solve the shortcomings from the previous course edition. In particular we wanted to avoid requiring dedicated GPUs for the RAP course using a model where GPUs would be shared with other courses.
To this end we installed a Kubernetes cluster on the same Openstack infrastructure, this hid physical and virtual hosts from students who were provided scripts to directly run "pods" (i.e., collections of containers) on the distributed cluster.
To enable GPU sharing, we used the \texttt{nvidia-docker} runtime which provides access to GPUs for containers running on a host -- any container running with this runtime will have access to the GPU: no fine-grained control over resources is however supported, meaning that any single container can consume all the resources of a single GPU. A quick empirical validation allowed us to estimate that running two RAP groups on the same GPU would not cause a perceivable performance decay, so we configured the system to allocate shared GPUs for maximum two RAP groups concurrently.
This meant that in 2022 we could support 24 RAP students (in 8 groups) with only 4 GPUs.

As noted above, one of the key drivers for this approach is to \emph{support sharing of GPUs to allow multiple robotic applications to leverage deep learning models}. 
Sharing nvidia GPUs in containerized environments is evolving with the release of Multi-instance GPUs (MIG)\footnote{\url{https://www.nvidia.com/en-us/technologies/multi-instance-gpu/}} which is  a promising solution which will support accurate control of GPU resources. Our approach, however, was to use a simpler solution based on technologies with which we already had experience.

The main drawback of using a K8S cluster is that we had to take special care with networking to the robots.
While TCP connectivity as required by ROS is generally possible by configuring the ingress-controller of a K8S cluster\footnote{\url{https://kubernetes.github.io/ingress-nginx/user-guide/exposing-tcp-udp-services/}}, the cluster was shared by multiple courses and was installed with a minimal configuration: only HTTPS traffic through a proxy was allowed. This is a fairly common restriction also in public cloud ``managed K8S cluster'' offerings.

\begin{figure}[thpb]
  \centering      
  \includegraphics[scale=0.71]{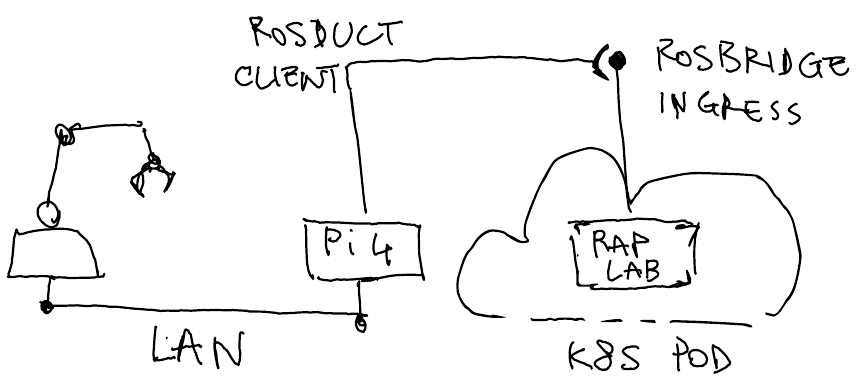}
  \caption{The robot and Pi use a rosduct client to relay ROS traffic to the K8S pod}
  \label{fig:rosbridge}
\end{figure}

To circumvent this issue, we resorted to using web sockets as done in \cite{fogros}: the \texttt{rosbridge} suite was running in the K8S pods and the robots would use \texttt{rosduct} to connect to it as in Figure \ref{fig:rosbridge}. We customized both components to support bidirectional CBOR (Concise Binary Object Representation). While this worked for the scope of our course, the solution can be further improved. We observed frequent unexpected disconnections from the websocket - it was reestablished quickly so it was not unusable when throttling message rates\footnote{An example of running the system to control a real Niryo arm is available at \url{https://www.youtube.com/watch?v=CWYd-MeHG6c}} but it led to some performance degradation. Also, there were still some issues with some specific CBOR data types encoding due to our quick implementation\footnote{\url{https://github.com/icclab/rosduct} and \url{https://github.com/icclab/rosbridge_suite/tree/ros1}}.

\subsection{Summer 2022: K8S and VPN sidecar}

\begin{figure}[thpb]
  \centering      
  \includegraphics[scale=0.71]{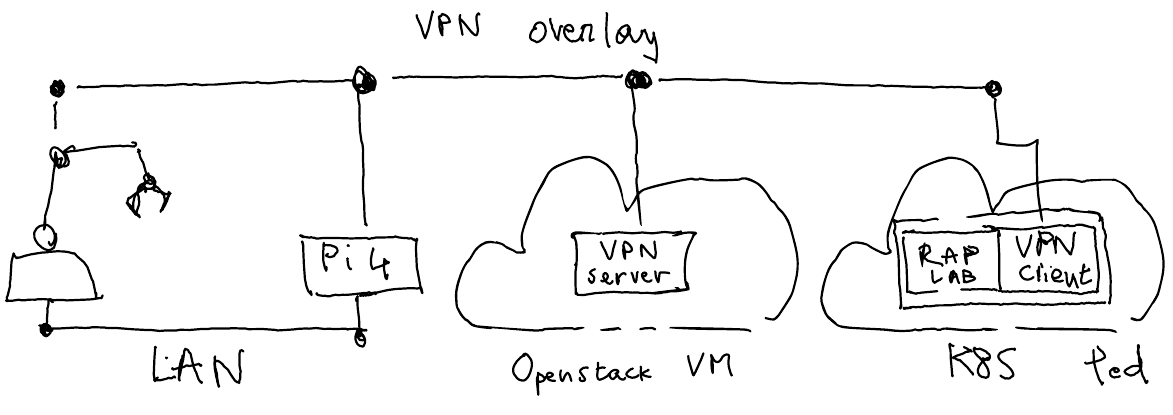}
  \caption{The VPN overlay built connecting robots and edge devices in an isolated LAN with a K8S pod through the sidecar pattern}
  \label{fig:vpn_niryo}
\end{figure}

By the end of the semester, with the need of controlling the Summit XL for the challenge and supporting its higher bandwidth requirements (i.e., streaming of two RGBD cameras and point clouds, two laser scanners) we needed to resolve the connectivity issues we had with web sockets.
For lack of a possibility to update the K8S cluster, we resorted to use an external VM as VPN server and run the ROS node network on top of a VPN overlay. This also required manual ROS environmental variables adjustment to mitigate the lack of DNS resolution.
While installing an OpenVPN client (either directly or containerized) on the robots was simple, we didn't want to rebuild and redistribute our RAP container images. Using the \emph{sidecar pattern} in the K8S pods allowed us to add a container that would create the VPN tunnel and make it available for the entire pod, granting a network interface our original container could use to access the VPN overlay. We ended up adapting a similar configuration we found online\footnote{\url{https://bugraoz93.medium.com/openvpn-client-in-a-pod-kubernetes-d3345c66b014}}.
A high level representation of the set up is depicted in Figure \ref{fig:vpn_niryo}.

With this configuration, we could support using a Web browser to fully control our Summit XL mobile manipulator in both navigation and grasping task through the cloud, including sending the point cloud from the arm camera to be used for segmentation and grasp generation with GPD.

\section{CLOUD NATIVE ROBOTIC APPS}

As Spring 2023 approaches, a new edition of the RAP course lingers, and this year we will have a dedicated K8S cluster on shared infrastructure. This allows sharing GPUs but removes the limitation on incoming TCP traffic, meaning we could host a VPN server for each student group, configuring all pods of a student group namespace to have access to the VPN overlay\footnote{See for example here: \url{https://docs.k8s-at-home.com/guides/pod-gateway/}}.
In this scenario, traffic would no longer be routed through an external VPN server. On top of shorter routing, in a public cloud deployment this would also mean \emph{not incurring in the additional charges of the provider's VPN services}.

\begin{figure}[thpb]
  \centering      
  \includegraphics[scale=0.71]{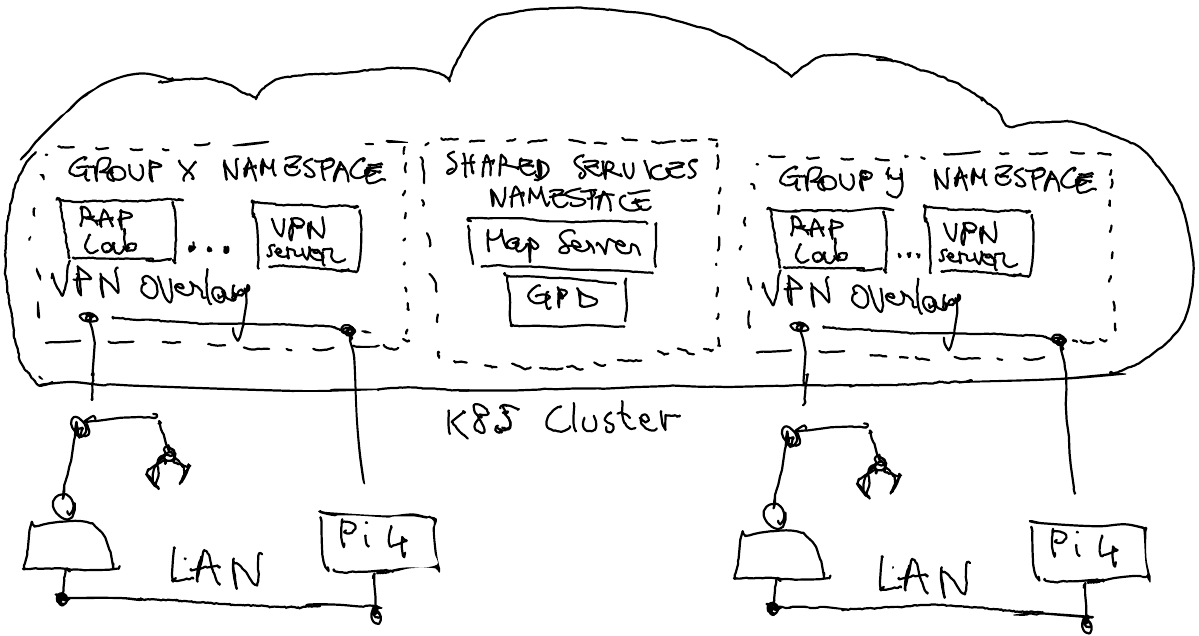}
  \caption{A more cloud-native deployment of a cloud-to-edge robotic application. Shared services and dynamic orchestration.}
  \label{fig:cloud_native}
\end{figure}

Given this setup it makes a lot more sense to rewrite our labs to enable \emph{sharing} of commonly used services (e.g., the GPD grasp pose generation) \emph{across student groups}. This is in line with \emph{cloud native application practices} where  functionalities (i.e., K8S ``services'') are scaled and load balanced through multiple instances of their implementations (i.e., pods), see an example in Figure \ref{fig:cloud_native}.

Apart from optimizing resource usage, this would allow us to bring to the lectures the concept of \emph{orchestration} of dynamic components based on robotic behavior. Students would be required to switch on and off pods in their namespace (or respectively nodes on the robots) depending on a state machine / behavior tree (e.g., stream the arm camera and process the stream only when an object to be picked is detected by the front camera). 
This would reduce even further our need of GPU resources, allowing us to host even more students per GPU, and would also teach students how to write applications that minimize energy and resource consumption.

\section{CONCLUSIONS}

In this paper we discussed our different setups and experiences in teaching a robotic application programming course that leverages containerized cloud computing resources connected to local robotic hardware at our university.
Notwithstanding our specific networking setup, the solutions we used in the past and we propose for the future are also applicable to a public cloud scenario with non publicly addressable robots (e.g., on a private LAN, behind NAT) and should be generally useful for other teachers and cloud robotics practitioners willing to share cloud GPU resources across robotic applications.

\addtolength{\textheight}{-19.5cm}   % This command serves to balance the column lengths

%%%%%%%%%%%%%%%%%%%%%%%%%%%%%%%%%%%%%%%%%%%%%%%%%%%%%%%%%%%%%%%%%%%%%%%%%%%%%%%%

%%%%%%%%%%%%%%%%%%%%%%%%%%%%%%%%%%%%%%%%%%%%%%%%%%%%%%%%%%%%%%%%%%%%%%%%%%%%%%%%



\section*{ACKNOWLEDGMENT}

This work is partially funded by the European Union under the NEPHELE project. This project has received funding from the European Union’s Horizon Europe research and innovation programme under grant agreement No 101070487. Views and opinions expressed are however those of the author(s) only and do not necessarily reflect those of the European Union. 

% The preferred spelling of the word ÒacknowledgmentÓ in America is without an ÒeÓ after the ÒgÓ. Avoid the stilted expression, ÒOne of us (R. B. G.) thanks . . .Ó  Instead, try ÒR. B. G. thanksÓ. Put sponsor acknowledgments in the unnumbered footnote on the first page.

%%%%%%%%%%%%%%%%%%%%%%%%%%%%%%%%%%%%%%%%%%%%%%%%%%%%%%%%%%%%%%%%%%%%%%%%%%%%%%%%

% References are important to the reader; therefore, each citation must be complete and correct. If at all possible, references should be commonly available publications.

\end{document}